\def\year{2016}
\documentclass[letterpaper]{article}
\usepackage{aaai}
\usepackage{url}
\usepackage{times}
\usepackage{epsfig}
\usepackage{graphicx}
\usepackage{subfig}
\usepackage{epstopdf}
\usepackage{amsmath}
\usepackage{amssymb}
\usepackage{enumitem}
\usepackage[ruled,noend, linesnumbered]{algorithm2e}
\let\oldnl\nl
\newcommand{\nonl}{\renewcommand{\nl}{\let\nl\oldnl}}

\usepackage{subscript}

\usepackage{siunitx} 
\usepackage{dsfont}
\usepackage[table,usenames]{xcolor}
\usepackage{helvet}
\usepackage{courier}

\usepackage{dcolumn}
\newcolumntype{d}[1]{D{.}{.}{#1} }

\usepackage{xspace}
\makeatletter
\DeclareRobustCommand\onedot{\futurelet\@let@token\@onedot}
\def\@onedot{\ifx\@let@token.\else.\null\fi\xspace}

\def\eg{\emph{e.g}\onedot}

\def\etc{\emph{etc}\onedot} 
\def\wrt{w.r.t\onedot} 

\makeatother
\newcommand{\st}{\mathrm{s.t.}}
\newcommand{\svec}{\mathbf{s}}
\newcommand{\avec}{\mathbf{a}}

\newcommand{\wvec}{\mathbf{w}}
\newcommand{\Wmat}{\mathbf{W}}

\newcommand{\Iit}{\mathit{I}}

\newcommand{\Ncal}{\mathfrak{N}}
\newcommand{\Pcal}{\mathfrak{P}}

\usepackage{cleveref}

\begin{document}
%
\title{Dynamic Concept Composition for Zero-Example Event Detection}
\author{
Xiaojun Chang$^1$, Yi Yang$^1$, Guodong Long$^1$, Chengqi Zhang$^1$~and~Alexander G. Hauptmann$^2$\\
$^1$Centre for Quantum Computation and Intelligent Systems, University of Technology Sydney. \\
$^2$Language Technologies Institute, Carnegie Mellon University. \\
\{cxj273, yee.i.yang\}@gmail.com, \{guodong.long,chengqi.zhang\}@uts.edu.au, alex@cs.cmu.edu
}

\maketitle
\begin{abstract}
\begin{quote}
In this paper, we focus on automatically detecting events in unconstrained videos without the use of any visual training exemplars. In principle, zero-shot learning makes it possible to train an event detection model based on the assumption that events (\eg \emph{birthday party}) can be described by multiple mid-level semantic concepts (\eg ``blowing candle'', ``birthday cake''). Towards this goal, we first pre-train a bundle of concept classifiers using data from other sources. Then we evaluate the semantic correlation of each concept \wrt the event of interest and pick up the relevant concept classifiers, which are applied on all test videos to get multiple prediction score vectors. While most existing systems combine the predictions of the concept classifiers with fixed weights, we propose to learn the optimal weights of the concept classifiers for each testing video by exploring a set of online available videos with free-form text descriptions of their content. To validate the effectiveness of the proposed approach, we have conducted extensive experiments on the latest TRECVID MEDTest 2014, MEDTest 2013 and CCV dataset. The experimental results confirm the superiority of the proposed approach.
\end{quote}
\end{abstract}

\section{Introduction}

In multimedia event detection (MED), a large number of \emph{unseen} videos is presented and the the learning algorithm must rank them according to their likelihood of containing an event of interest, such as \emph{rock climbing} or \emph{attempting a bike trick}. Compared to traditional recognition of visual concepts (\eg actions, scenes, objects, \etc.), event detection is more challenging for the following reasons. First, an event is a higher level abstraction of video sequences than a concept and consists of multiple concepts. For example, an event, say \emph{birthday part}, can be described by multiple concepts (\eg ``birthday cake'', ``blowing candle'', \etc.) Second, an event spreads over the entire duration of long videos while a concept can be detected in a shorter video sequence or even in a single frame. As the first important step towards automatic categorization, recognition, search, indexing and retrieval, MED has attracted more and more research attention in the computer vision and multimedia communities~\cite{ChenCYLC14,ChangYXY15,ChengFPC14,LaiYCC14,LiYDV13,ChangYYH15,MaYCSH12,YanYMLTHS15,YanYSMLHS15}.

Current state-of-the-art systems for event detection first seek a compact representation of the video using feature extraction and encoding with a pre-trained codebook \cite{Lowe04,BayTG06,WangS13a}. With labeled training data, sophisticated statistical classifiers, such as support vector machines (SVM), are then applied on top to yield predictions. With sufficient labeled training examples, these systems have achieved remarkable performance in the past \cite{LaiYCC14,SunN14,LiYDV13,ChengFPC14}. However, MED faces the severe data-scarcity challenge: only very few, perhaps even none, positive training samples are available for some events, and the performance degrades dramatically once the number of training samples falls short. Reflecting this challenge, the National Institute of Standards and Technology (NIST) hosts an annual competition on a variety of retrieval tasks, of which the Zero-Exemplar Multimedia Event Detection (0Ex MED in short) in TRECVID 2013 \cite{MED13} and 2014 \cite{MED14} has received considerable attention. Promising progress \cite{DaltonAM13,HabibianSS13,HabibianMS14,ChangYHXY15} has been made in this direction, but further improvement is still anticipated.

In this paper, we aim to detect complex event without any labeled training data for the event of interest. Following previous work on zero-shot learning \cite{LampertNH09,PalatucciPHM09}, we regard an event as compositions of multiple mid-level semantic concepts. These semantic concept classifiers are shared among events and can be trained using other resources. We then learn a skip-gram model \cite{MikolovSCCD13} to assess the semantic correlation of the event description and the pre-trained vocabulary of concepts, based on which we automatically select the most relevant concepts to each event of interest. This step is carried out without any visual training data at all. Such concept bundle view of event also aligns with the cognitive science literature, where humans are found to conceive objects as bundles of attributes \cite{RoahchL78}. The concept prediction scores on the testing videos are combined to obtain a final ranking of the presence of the event of interest. However, most existing zero-shot event detection systems aggregate the prediction scores of the concept classifiers with fixed weights. Obviously, this assumes all the predictions of a concept classifier share the same weight and fails to consider the differences of the classifier's prediction capability on individual testing videos. A concept classifier, in fact, does have different prediction capability on different testing videos, where some videos are correctly predicted while others are not. Therefore, instead of using a fixed weight for each concept classifier, a promising alternative is to estimate the specific weight for each testing video to alleviate the individual prediction errors from the imperfect concept classifiers and achieve robust detection result.

The problem of learning specific weights of all the semantic concept classifiers for each testing video is challenging in the following aspects: Firstly, it is unclear how to determine the specific weights for the unlabeled testing video since no label information can be used. Secondly, to get a robust detection result, we need to maximally ensure positive videos have higher scores than negative videos in the final ranking list. Note that the goal of event detection is to rank the positive videos above negative ones. To this end, we propose to learn the optimal weights for each testing video by exploring a set of online available videos with free-form text descriptions of their content. Meanwhile, we directly enforce that positive testing videos have the highest aggregated scores in the final result.

The main building blocks of the proposed approach for zero-example event detection can be described as follows. We first rank the semantic concepts for each event of interest using the skip-gram model, based on which the relevant concept classifiers are selected. Then following \cite{LiuLYCC13,LaiLCC15}, we define the aggregation process as an information propagation procedure which propagates the weights learned on individual on-line available videos to the individual unlabeled testing videos, which enforces visually similar videos have similar aggregated scores and offers the capability to infer weights for the testing videos. To step further, we use the $L_{\infty}$ norm infinite push constraint to minimize the number of positive videos ranked below the highest-scored negative videos, which ensures most positive videos have higher aggregated scores than negative videos. In this way, we learn the optimal weights for each testing video and push positive videos to rank above negative videos as possible. 

\textbf{Contributions}: To summarize, we make the following contributions in this work:

\begin{enumerate}
	\item We propose a novel approach for zero example event detection to learn the optimal weights of related concept classifiers for each testing video by exploring a set of online available videos with free-form text descriptions of their content.
	\item Infinity push SVM has been incorporated to ensure most positive videos have the highest aggregated scores in the final prediction results.
	\item We conduct extensive experiments on three real video datasets (namely MEDTest 2014 dataset, MEDTest 2013 dataset and CCV\textsubscript{sub}), and achieve state-of-the-art performances.
\end{enumerate}

\section{Related Works}
\label{sec:related}

Complex event detection on unconstrained web videos has attracted wide attention in the field of multimedia and computer vision. Significant progress has been made in the past \cite{LaiYCC14,LiYDV13,SunN14}. A decent video event detection system usually consists of a good feature extraction module and a highly effective classification module (such as large margin support machines and kernel methods). Various low-level features (static, audio, \etc.) already achieve good performances under the bag-of-words representation. Further improvements are obtained by aggregating complementary features in the video level, such as coding \cite{BoureauBLP10,PerronninSM10} and pooling \cite{CaoMNCHS12}. It is observed that with enough labeled training data, superb performance can be obtained. However, when the number of positive training videos falls short, the detection performance drops dramatically. In this work, we focus on the more challenging zero-exemplar setting where \emph{no} labeled training videos for the event of interest are provided.

Our work is inspired by the general zero-shot learning framework \cite{LampertNH09,PalatucciPHM09,MensinkGS14}, which arises from practical considerations such as the tremendous cost of acquiring labeled data and the constant need of dealing with  dynamic and evolving real-world object categories. On the event detection side, recent works have begun to explore intermediate semantic concepts \cite{ChangYHXY15}, and achieved limited success on the zero-exemplar setting \cite{DaltonAM13,HabibianSS13,HabibianMS14} also considered selecting more informative concepts. However, none of these works consider discovering the optimal weights of different concept classifiers for each \emph{individual} testing video.

\section{The Proposed Approach}
\label{sec:approach}

In this paper, we focus on the challenging zero-exemplar event detection problem. In a nutshell, we are given a sequence of unseen testing videos and also the event description, but without any labeled training data for the event of interest. The goal is to rank the testing videos so that positive videos (those contain the event of interest) are ranked above negatives. With this goal in mind, we first associate a query event with related semantic concepts that are pre-trained using other sources. Then we aggregate the individual concept prediction scores using the proposed dynamic composition approach.

\subsection{Semantic Query Generation}
\label{sec:sqg}

Our work is built upon the observation that each event can be described by multiple \emph{semantic concepts}. For example, the \emph{marriage proposal} event can be attributed to several concepts, such as ``ring'' (object), ``kissing'' (action), ``kneeling down'' (action) and ``cheering'' (acoustic). Since semantic concepts are shared among different events and each concept classifier can be trained independently using data from other sources, zero-example event detection can be achieved by combining the relevant concept prediction scores. Different from the pioneer work \cite{LampertNH09}, which largely relies on human knowledge to decompose classes (events) into attributes (concepts), our goal is to automatically evaluate the semantic similarity between the event of interest and the concepts, based on which we select the relevant concepts for each event.

Events come with textual side informatin, \eg, an event name or a short description. For example, the event \emph{dog show} in the TRECVID MEDTest 2014 \cite{MED14} is defined as ``a competitive exhibition of dogs''. With the availability of a pre-trained vocabulary of concept classifiers, we can evaluate the semantic correlation between the query event and each individual concepts. Specifically, we learn a skip-gram model \cite{MikolovSCCD13} using the English Wikipedia dump\footnote{http://dumps.wikimedia.org/enwiki/}. The skip-gram model infers a $D$-dimensional vector space representation by fitting the joint probability of the co-occurrence of surrounding contexts on large unstructured text data, and places semantically similar words near each other in the embedding vector space. Thus it is able to capture a large number of precise syntactic and semantic word relationships. For short phases consisting of multiple words (\eg, event descriptions), we simply average its word-vector representations. After properly normalizing the respective word-vectors, we compute the cosine distance of the event description and all individual concepts, resulting in a correlation vector $\mathbf{w} \in [0,1]^m$, where $w_k$ measures a priori relevance of the $k$-th concept and the event of interest. Based on the relevance vector, we select the most informative concept classifiers for each event.

\subsection{Weak Label Generation}
\label{sec:wlg}

According to the NIST standard, we utilize the TRECVID MED \emph{research} dataset to explore the optimal weights for the testing videos. All the videos in the \emph{research} set come with a sentence of description, summarizing the contents contained in the videos. Note that all the videos in the \emph{research} set has no event-level label information. We further collect a set of videos with free-form text descriptions of their content, which are widely available online in websites such as YouTube and NetFlix. 

As the descriptions of the videos in both \emph{research} set and online website are very noisy, we apply standard natural language processing (NLP) techniques to clean up the annotations, including removal of the common stop words and stemming to normalize word inflections.

Similar to the steps of SQG, we measure the semantic correlation between the cleaned sentence and each concept description, and use it as a weak label for each concept. In the next section, we will learn the optimal aggregation weights for the individual testing video by exploiting the supervision information with weak label, which accounts for the differences in the concept classifiers' prediction abilities on the individual testing video, and hence achieve robust aggregation results.

\subsection{Dynamic Composition}
\label{sec:dc}

Up until now, we have $l$ videos with weak label and $u$ testing videos. We propose to learn an aggregation function $f_i(\svec_i) = \wvec_i^T \svec_i$ for each testing video ($i = 1, \cdots , l + u$), where $\wvec = [w_i^1, \cdots, w_i^m]^T$ is a non-negative aggregation weight vector with $w_i^j$ being the aggregation weight of $s_i^j$. Clearly, it is straightforward for us to learn the optimal aggregation weights for the videos in the collected set based on the weak label information. However, it is challenging to derive the optimal aggregation weights for the \emph{unlabeled} videos since no label information is available.  

To achieve this goal, we build our model based on the local smoothness property in graph-based semi-supervised learning, which assumes visually similar videos have comparable labels within a local region of the sample space \cite{LiuLYCC13,LaiLCC15}. Exploring the local connectivity of data is a successful strategy for graph construction. The neighbors of video $x_i$ can be defined as the $k$-nearest videos in the collection to $x_i$. In this work, we consider the probabilistic neighbors. Following the work in \cite{NieWH14}, we learn the data similarity matrix by assigning the adaptive neighbors for each video based on the local connectivity. 

For the $i$-th video $x_i$, all the videos $\{x_1, x_2, \dots, x_{l+u}\}$ can be connected to $x_i$ as a neighbor with probability $\avec_{ij}$. Usually, if $(\wvec_i^T\svec_i - \wvec_j^T\svec_j)^2$ is smaller, a larger probability $\avec_{ij}$ should be assigned between the video $x_i$ and $x_j$. So a natural method to determine the probabilities $\avec_{ij}|_{j=1}^{l+u}$ is solving the following problem:

\begin{align}
	\min_{\Wmat,\avec_i^T\mathbf{1} = 1, 0 \leq \avec_i \leq 1} \sum_{i,j=1}^{l+u} (\wvec_i^T\svec_i - \wvec_j^T\svec_j)^2\avec_{ij},
	\label{eq:trivial}
\end{align}
where $\svec_i$ is a vector with the $j$-th element as $\svec_{ij}$.

However, the problem \Cref{eq:trivial} has a trivial solution, only the nearest video can be the neighbor of the video $x_i$ with probability 1 and all the other videos can not be the neighbors of $x_i$. On the other hand, if the following problem is solved without involving any distance information between the videos:

\begin{align}
	\min_{\avec_i^T\mathbf{1} = 1, 0 \leq \avec_i \leq 1} \sum_{j=1}^{l+u} \avec_{ij}^2,
	\label{eq:prior}
\end{align}
the optimal solution is that all the data points can be the neighbors of the video $x_i$ with the same probability $\frac{1}{l+u}$, which can be seen as a prior in the neighbor assignment.

Combining \Cref{eq:trivial} and \Cref{eq:prior}, we can solve the following problem:
\begin{align}
	\min_{\Wmat,\avec_i^T\mathbf{1} = 1, 0 \leq \avec_i \leq 1} \sum_{j=1}^{l+u} (\wvec_i^T\svec_i - \wvec_j^T\svec_j)^2\avec_{ij} + \gamma \avec_{ij}^2.
	\label{eq:obj1}
\end{align}

The second term in \Cref{eq:obj1} is a regularization and $\gamma$ is the regularization parameter. Denote $d_{ij}^s = (\wvec_i^T\svec_i - \wvec_j^T\svec_j)^2$, the problem \Cref{eq:obj1} can be written in vector form as:

\begin{align}
	\min_{\avec_i^T\mathbf{1} = 1, 0 \leq \avec_i \leq 1} \| \avec_i + \frac{1}{2\gamma} d_{i}^s \|_2^2.
\end{align}
It is easy to verify this problem can be solved with a closed form solution.

To step further, we incorporate an infinite push loss function \cite{Rakotomamonjy12} to achieve robust aggregation result. The goal of infinite push loss function is to minimize the number of positive videos which are ranked below the highest scored negative videos. The infinite push loss function has shown promising performance for event detection problem in \cite{ChangYYH15}. In fact, the number of positive videos ranked below the highest scored negative videos equals to the maximum number of positive videos ranked below any negative videos. Hence, we define it as follows:

\begin{align}
	\label{eq:infinite}
	\ell(\{f_i\}_{i=1}^l; \Pcal, \Ncal) = \max_{j \in \Ncal}(\frac{1}{p} \sum_{i \in \Pcal}\Iit_{f_i(\svec_i^+) < f_j(\svec_j^-)}),
\end{align}
where $\Iit$ is the indicator function whose value is 1 if $f_i(\svec_i^+) < f_j(\svec_j^-)$ and 0 otherwise. The maximum operator over $j$ equals to calculating the $l_{\infty}$-norm of a vector consisting of $n$ entries, each of which corresponds to one value based on $j$ in the parentheses of \Cref{eq:infinite}. By minimizing this penalty, positive videos tend to score higher than any negative videos. This essentially ensures positive videos have higher combined scores than the negatives, leading to more accurate combined results.

For computational tractability we upper bound the discrete 0-1 loss $\Iit(\delta < 0)$ by the \emph{convex} hinge loss $(1 - \delta)_+$, where as usual $(\delta)_+ := \max(\delta,0)$ is the positive part. Since we usually par more attention, if not exclusively, to the top of the rank list, we focus on minimizing the maximum ranking error among all negative exemplars $j \in \Ncal$:

\begin{align}
	\ell (\{f_i\}_{i=1}^l; \Pcal, \Ncal) = \max_{j \in \Ncal} (\frac{1}{p}\sum_{i \in \Pcal}(1 - (\wvec_i^T\svec_i^+ - \wvec_j^T\svec_j^-))_+),
\end{align}

Finally, the objective function can be written as:

\begin{align}
	\label{eq:obj}
	&	\nonumber \min_{\Wmat,\avec_i^T\mathbf{1} = 1, 0 \leq \avec_i \leq 1} \sum_{j=1}^{l+u} (\wvec_i^T\svec_i - \wvec_j^T\svec_j)^2\avec_{ij} + \gamma \avec_{ij}^2 \\
	&	+ \lambda \max_{j \in \Ncal} (\frac{1}{p}\sum_{i \in \Pcal}(1 - (\wvec_i^T\svec_i^+ - \wvec_j^T\svec_j^-))_+), \\
	& \nonumber \st   \wvec_i \geq 0, i = 1, \cdots, l + u.
\end{align}

Since we are working on large-scale video event detection, there is a great challenge in the computation cost. Thanks to the proximal map \cite{Yu13}, we employ the faster ADMM proposed in \cite{ChangYYH15} for efficient solution. For space limitation, we omit the detailed optimization and will include in our journal version.

\section{Experiments}

In this section, we conduct extensive experiments to validate the proposed \textbf{D}ynamic \textbf{C}oncept \textbf{C}omposition for zero-exemplar event detection task, abbreviated as \textbf{DCC}.

\subsection{Experiment Setup}

\textbf{Dataset}: To evaluate the effectiveness of the proposed approach, we conduct extensive experiments on the following three large-scale event detection datasets:
\begin{enumerate}
	\item[---] TRECVID MEDTest 2014 dataset \cite{MED14}: This dataset has been introduced by the NIST for all participants in the TRECVID competition and research community to perform experiments on. There are in total 20 events, whose description can be found in \cite{MED14}. We use the official test split released by the NIST, and strictly follow its standard procedure \cite{MED14}. To be more specific, we detect each event \emph{separately}, treating each of them as a binary classification/ranking problem.
	\item[---] TRECVID MEDTest 2013 dataset \cite{MED13}: The settings of MEDTest 2013 dataset is similar to MEDTest 2014, with 10 of their 20 events overlapping.
	\item[---] Columbia Consumer Video dataset \cite{JiangYCEL11}: The official Columbia Consumer Video dataset contains 9,317 videos in 20 different categories, including scenes like ``beach'', objects like ``cat'', and events like ``basketball'' and ``parade''. Since the goal of this work is to \emph{search complex events}, we only use the 15 event categories.
\end{enumerate}

\begin{table*}[!ht]
	\vspace{-.5em}	
	\caption{Experiment results for 0Ex event detection on  MEDTest 2014, MEDTest 2013, and CCV\textsubscript{Sub}. Mean average precision (mAP), in percentages, is used as the evaluation metric. Larger mAP indicates better performance.}
	\vspace{-.5em}
	\label{tab:event_detection14}	
	\parbox{.08\linewidth}{
		\centering
		\textcolor[rgb]{1,1,1}{abcd}  \\
		\begin{tabular}{|>{\columncolor[gray]{0.9}}c|}
			\hline
			ID  \\
			\hline\hline 
			E021 \\
			\hline
			E022 \\
			\hline
			E023 \\
			\hline
			E024 \\
			\hline
			E025 \\
			\hline
			E026 \\
			\hline
			E027 \\
			\hline
			E028 \\
			\hline
			E029 \\
			\hline
			E030 \\
			\hline
			E031 \\
			\hline
			E032 \\
			\hline
			E033 \\
			\hline
			E034 \\
			\hline
			E035 \\
			\hline
			E036 \\
			\hline
			E037 \\
			\hline
			E038 \\
			\hline
			E039 \\
			\hline
			E040 \\
			\hline
			\hline
			mean \\
			\hline
			\hline
			\rowcolor{white} \\
			\hline
			\hline
			mean \\
			\hline
			\hline
			\rowcolor{white} \\
			\hline
			\hline
			mean \\
			\hline			
		\end{tabular}
	}
	\centering
	\parbox{0.91\linewidth}{
		\centering
		\rowcolors{2}{cyan!20}{pink!40}
		MEDTest 2014 \\
		\begin{tabular}{d{3.5}| d{3.5}| d{3.5}| d{3.5}| d{3.5}| d{3.5}| d{3.5}| d{3.5}}
			\hline
			\multicolumn{1}{c|}{Prim} & \multicolumn{1}{c|}{Sel}  & \multicolumn{1}{c|}{Bi} & \multicolumn{1}{c|}{OR} & \multicolumn{1}{c|}{Fu} & \multicolumn{1}{c|}{Bor} & \multicolumn{1}{c|}{PCF} & \multicolumn{1}{c|}{DCC}  \\
			\hline\hline
			2.12 & 2.98 & 2.64    & 3.89 & 3.97 &  3.12  & 4.64 & 6.37  \\
			\hline
			0.75  & 0.97 & 0.83   & 1.36  & 1.49  & 1.15  & 1.48 & 2.85   \\
			\hline
			33.86 & 36.94 & 35.23   & 39.18 & 40.87  & 38.68 & 41.78 & 44.26 \\
			\hline
			2.64   & 3.75 & 3.02   & 4.66 & 4.92  &  4.11 & 4.87 & 6.12  \\
			\hline
			0.54  & 0.76 & 0.62   & 0.97 &  1.39  &  0.84  & 1.01 & 1.26  \\
			\hline
			0.96    & 1.59  & 1.32   & 2.41 &  2.96  & 1.96  & 2.65 & 4.23  \\
			\hline
			11.21   & 13.64  & 12.48   & 15.93 &  16.26  & 15.12 & 16.47 & 19.63  \\
			\hline
			0.79   & 0.67  & 1.06  & 1.57  &  1.95  &  1.72  & 2.25 & 4.04  \\
			\hline
			8.43  & 10.68  & 12.21  & 14.01  &  14.85  & 13.19  & 14.75  & 17.69  \\
			\hline
			0.35   & 0.63  & 0.48 & 0.91 & 0.96  & 0.36 & 0.48 & 0.52  \\
			\hline
			32.78   & 53.19 & 45.87   & 69.52 & 69.66  &  67.49  & 72.64 & 77.45  \\
			\hline
			3.12   & 5.88 & 4.37   & 8.12 &  8.45   & 7.54  & 8.65 & 11.38   \\
			\hline
			15.25    & 20.19 & 18.54   & 22.14  &  22.23  &   21.53  & 23.26 & 26.64  \\
			\hline
			0.28   & 0.47 & 0.41   & 0.71 &  0.75  & 0.53  & 0.76 & 0.94   \\
			\hline
			9.26    & 13.28 & 11.09   & 16.53  &  16.68 &   15.82  & 18.65 & 21.78   \\
			\hline
			1.87    & 2.63 & 2.14   & 3.15  & 3.39  & 2.88  & 3.76 & 5.47   \\
			\hline
			2.16   & 4.52 & 3.81    & 6.84  & 6.88  &  5.42  & 6.83 & 8.45   \\
			\hline
			0.66   & 0.74 & 0.58    & 0.99 & 1.16  &  0.85  & 1.12 & 2.89  \\
			\hline
			0.36  & 0.57 & 0.42    & 0.69 &  0.77  &  0.64  & 0.85 & 2.26  \\
			\hline
			0.65    & 0.98 & 0.72   & 1.57 &  1.57  &   1.24  & 1.76 & 3.12  \\
			\hline \hline
			\rowcolor{white}
			6.40  & 9.55  & 7.89 & 10.76  &  11.05 &   10.21 & 11.44 & 13.37   \\
			\hline\hline
			\rowcolor{white}
			\multicolumn{8}{c}{MEDTest 2013}\\			
			\hline\hline
			\rowcolor{white}
			7.07  & 7.94  & 6.92 & 9.45  &  9.88 &  8.43 & 9.96 & 12.64   \\
			\hline\hline
			\rowcolor{white}
			\multicolumn{8}{c}{CCV\textsubscript{sub}}\\			
			\hline\hline
			\rowcolor{white}
			19.05  & 19.40  & 20.25 &   21.16  &  21.89 &  23.08 & 23.87 & 24.36   \\
			\hline			
		\end{tabular}
	}
	\vspace{-.8em}
\end{table*}

According to the standard of the NIST, each event is detected separately and the performance of event detection is evaluated using the mean Average Precision (mAP).

\textbf{Concept Detectors}: 3,135 concept detectors are pre-trained using TRECVID SIN dataset (346 categories) \cite{OAMFSKSQ2014,JiangMYLSH14}, Google sports (478 categories) \cite{KarpathyTSLSF14,JiangMYLSH14}, UCF101 dataset (101 categories) \cite{SoomroZS12,JiangMYLSH14}, YFCC dataset (609 categories) \cite{YFCC,JiangMYLSH14} and DIY dataset (1601 categories) \cite{YuJH14,JiangMYLSH14}. The improved dense trajectory features (including trajectory, HOG, HOF and MBH) are first extracted using the code of \cite{WangS13a} and encode them with the Fisher vector representation \cite{PerronninSM10}. Following \cite{WangS13a}, the dimension of each descriptor is first reduced by a factor of 2 and then use 256 components to generate the Fisher vectors. Then, on top of the extracted low-level features, the cascade SVM \cite{GrafCBDV04} is trained for each concept detector. 

\textbf{Competitors}: We compare the proposed approach with the following alternatives: 1). Prim \cite{HabibianMS14}: Primitive concepts, separately trained. 2). Sel \cite{MazloomGSS13}: A subset of primitive concepts that are more informative for each event. 3). Bi \cite{RastegariDPF13}: Bi-concepts discovered in \cite{RastegariDPF13}. 4). OR \cite{HabibianMS14}: Boolean OR combinations of Prim concepts. 5). Fu \cite{HabibianMS14}: Boolean AND/OR combinations of Prim concepts, w/o concept refinement. 6). Bor: The Borda rank aggregation with equal weights on the discovered semantic concepts. 7). PCF \cite{ChangYHXY15}: The pair-comparison framework is incorporated for zero-shot event detection.

\subsection{Zero-exemplar event detection}

\begin{figure*}[th]
	\vspace{-.5em}
	\centering
	\includegraphics[scale=1.]{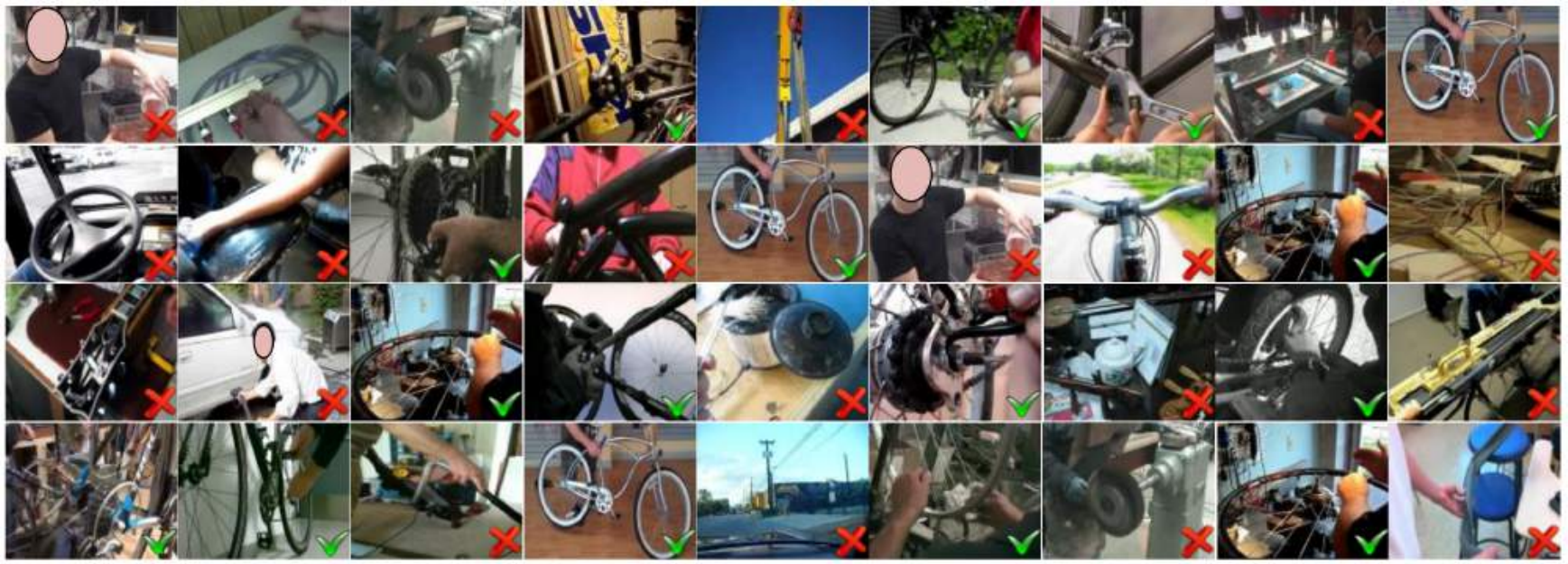}
	\caption{Top ranked videos for the event \emph{non-motorized vehicle repair}. From top to below: OR, Fu, PCF and DCC. True/false labels (provided by NIST) are marked in the lower-right of each frame.}
	\label{fig:top_ranked}
\end{figure*}

We report the full experimental results on the TRECVID MEDTest 2014 dataset in \Cref{tab:event_detection14} and also a summary on the MEDTest 2013 dataset and CCV\textsubscript{sub}. From the experimental results shown in \Cref{tab:event_detection14}, the proposed algorithm, \textbf{DCC}, performs better than the other approaches with a large margin (13.37\% vs 11.44\% achieved by PCF). The proposed approach gets significant improvement on some vents, such as \emph{Dog Show (E023)}, \emph{Rock Climbing (E027)}, \emph{Beekeeping (E031)} and \emph{Non-motorized Vehicle Repair (E033)}. By analyzing the discovered concepts for these events, we find that their classifiers are very discriminative and reliable. For example, for the event \emph{Rock Climbing}, we discovered the concepts ``Sport climbing'', ``Person climbing'' and ``Bouldering'', which are the most informative concepts for \emph{Rock climbing} in the concept vocabulary. \Cref{fig:top_ranked} illustrates the top retrieved results on the \emph{non-motorized vehicle repair} event. To save space, we only show the top 4 compared algorithms (OR, Fu, PCF and DCC).  It is clear that videos retrieved by the proposed DCC are more accurate and visually coherent.

We make the following observations from the results shown in \Cref{tab:event_detection14}: 1). Sel significantly outperform Prim with mAP of 9.55\% vs 6.40\% on MEDTest 2014, which indicates that selecting the most discriminative concepts generally improves detection performance than naively using all the concepts. 2). Comparing the results of Bi, OR, Fu and Bor, we verify that the selected informative concept classifiers are not equally important for event detection. It is beneficial to derivate the weights for each concept classifier. By treating the concept classifiers differentially, better performance is achieved. 3). Comparing the results of DCC with the other alternatives, we observe that learning optimal weights of concept classifiers for each testing video significantly improves performance of event detection. This confirms the importance of dynamic concept composition.

We made similar observations from the results on the TRECVID MEDTest 2013 dataset and CCV\textsubscript{sub} dataset.

\subsection{Extension to few-exemplar event detection}

\label{extension}
\begin{figure*}[!th]
	\vspace{-1.0em}
	\centering
	\includegraphics[scale=1.]{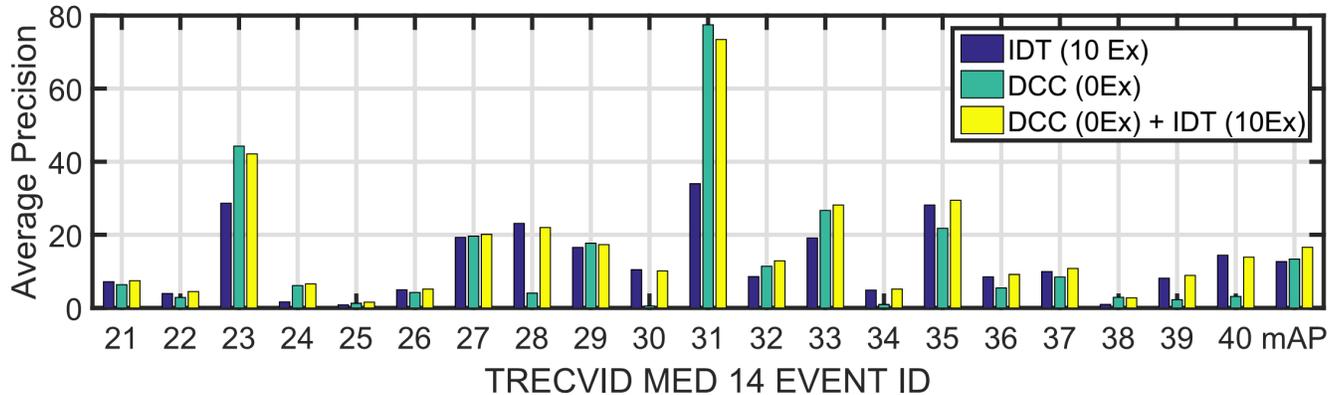}
	\caption{Performance comparison of IDT, DCC, and the hybrid of IDT and DCC.}
	\vspace{-1.0em}
	\label{fig:extension_med14}
\end{figure*}

The proposed zero-example event detection framework can also be used for few-exemplar event detection: we aggregate the concept classifiers and the supervised classifier using the dynamic concept composition approach. In this section, experiments are conducted to demonstrate the benefit of this hybrid approach. \Cref{tab:extension} summarizes the mAP on both the MEDTest 2014 and 2013 datasets, while \Cref{fig:extension_med14} compares the performance event-wise.

According to the NIST standard, we consider the 10 Ex setting, where 10 positive videos are given for each event of interest. The improved dense trajectories feature \cite{WangS13a} is extracted, on top of which an SVM classifier is trained. It is worthwhile to note that our DCC which had no labeled training data can get comparable results with the supervised classifier (mAP 13.37\% vs 13.92\% on MEDTest 2014). This demonstrates that proper aggregation with optimal weights for each testing video can get promising results for event detection.

\begin{table}
	\footnotesize	
	\centering
	\caption{Few-exemplar results on MED14 and MED13.}
	\vspace{-.8em}
	\label{tab:extension}	
	\begin{tabular}{c||c|c}
		\hline
		DCC (0Ex) & 13.37 & 12.64 \\
		\hline
		IDT (10Ex) & 13.92 &    18.08   \\
		\hline
		DCC (0Ex) + IDT (10Ex) & 16.37 &  19.24  \\
		\hline		
	\end{tabular}
	\vspace{-1.5em}	
\end{table}

We compare DCC with IDT event-wise in \Cref{fig:extension_med14}. From the results we can see that our DCC outperforms the supervised IDT on multiple events, namely E023, E024, E031 and E033. To be specific, on the event \emph{Beekeeping} (E031), DCC significantly outperform the supervised IDT (77.45\% vs 33.92\%). This is not supervising, since DCC significantly benefited from the presence of informative and reliable concepts such as ``apiary bee house'' and ``honeycomb'' on the particular  \emph{Beekeeping} event. 


Finally we combine DCC with IDT to get the final few-example event detection result. This improves the performance from 13.92\% to 16.98\% on the MEDTest 2014 dataset. As expected, the gain obtained from such simple hybrid diminishes when combining with more sophisticated methods. 
Overall, the results clearly demonstrate the utility of our framework even in the few-exemplar setting.

\vspace{-1em}
\section{Conclusions}
\label{sec:conclusion}

To address the challenging task of zero-exemplar or few-exemplar event detection, we proposed to learn the optimal weights for each testing video by exploring the collected videos from other sources. Data-driven word embedding models were used to seek the relevance of the concepts to the event of interest. To further derivate the optimal weights of the concept classifiers for each testing video, we have proposed a novel dynamic concept composition method by exploiting the textual information of the collected videos. Extensive experiments are conducted on three real video datasets. The experimental results confirm the efficiency of the proposed approach.

\section*{Acknowledgment}
This paper was partially supported by the ARC DECRA project DE130101311, partially supported by the ARC discovery project DP150103008, partially supported by the US Department of Defense, U. S. Army Research Office (W911NF-13-1-0277) and by the National Science Foundation under Grant No. IIS-1251187. The U.S. Government is authorized to reproduce and distribute reprints for Governmental purposes notwithstanding any copyright annotation thereon.

{
\small
\bibliographystyle{aaai}
\bibliography{zero_shot}
}

\end{document}